\documentclass[letterpaper, 10 pt, conference]{ieeeconf}  

\IEEEoverridecommandlockouts                              

\usepackage{graphics} 
\usepackage{epsfig} 
\usepackage{times} 
\usepackage{amsmath} 
\usepackage{amssymb}  
\usepackage{booktabs}
\usepackage{multirow}  
\usepackage{makecell}  
\usepackage{caption}
\usepackage{subcaption}
\usepackage{algorithm}
\usepackage{algpseudocode}
\usepackage{xcolor}

\title{\LARGE \bf
Co-VLA: Coordination-Aware Structured Action Modeling for Dual-Arm Vision–Language–Action Systems
}
\author{Yandong Wang$^{1,2}$, Jiaqian Yu$^{2}$, Xiongfeng Peng$^{2}$, Lu Xu$^{2}$, Yamin Mao$^{2}$, Weiming Li$^{2}$, \\
Jaewook Yoo$^{3}$, Dongwook Lee$^{3}$, Daehyun Ji$^{3}$, Mingbo Zhao$^{1}$, Chao Zhang$^{2}$
\thanks{$^{1}$Yandong Wang and Mingbo Zhao is with Donghua University, Shanghai, China}%
\thanks{$^{2}$Yandong Wang, Jiaqian Yu, Xiongfeng Peng, Lu Xu, Yamin Mao, Weiming Li, and Chao Zhang are with Samsung R\&D Institute China-Beijing (SRCB), China}%
\thanks{$^{3}$Jaewook Yoo, Dongwook Lee, and Daehyun Ji are with Samsung AI Center, DS Division, South Korea}%
}

\begin{document}
\IEEEaftertitletext{
\begin{center} 
    \centering \includegraphics[width=0.99\linewidth]{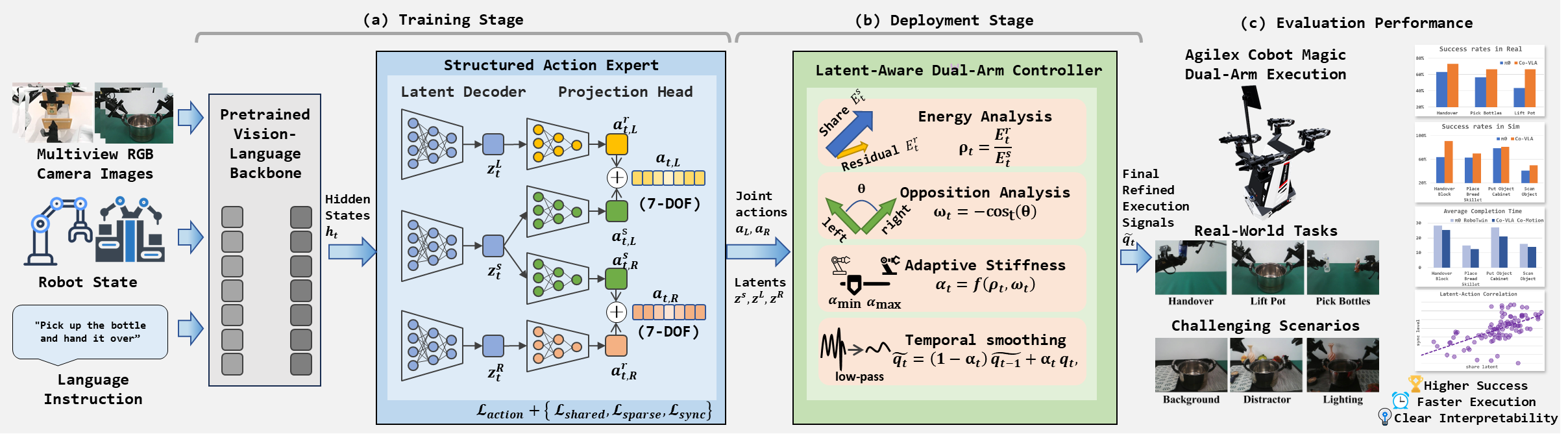}
    \captionof{figure}{Overview of our dual-arm VLA system.
We design a Structured Action Expert (SAE) that interprets the hidden states from the VLM backbone into a structured latent space, then produces decomposed actions through separate projection heads. 
At deployment time, a Latent-Aware Controller (LAC) interprets the learned coordination representations to refine joint-level commands for accurate, synchronized, smooth, and safe execution. }
    \label{fig:system_overview}
\end{center}%
}
\maketitle 
\thispagestyle{empty}
\pagestyle{empty}

\begin{abstract}
Vision–language–action (VLA) models have recently demonstrated strong capabilities in both single-arm and dual-arm robotic manipulation. Prior works have shown that coordinated bimanual behaviors can emerge from end-to-end learning frameworks, highlighting the expressive power of large vision–language backbones combined with continuous action prediction. 
However, as bimanual tasks become more tightly coupled and execution constraints become more critical, implicit coordination alone may be insufficient to ensure reliable, interpretable, and deployment-stable behavior.
In this work, we propose \textbf{Co-VLA}, a coordination-aware bimanual manipulation framework that introduces explicit structural priors into VLA models. Specifically, we instantiate our method on a state-of-the-art vision-language backbone by replacing its monolithic action head with a \textbf{Structured Action Expert (SAE)} explicitly designed for bimanual coordination. 
In particular, we introduce explicit structure at the action generation level together with a modular coordination-aware loss design that shapes shared and residual latents according to task-specific bimanual coordination structures. The shared latent encodes task-level coordination intent, while residual latents capture residual execution adjustments for each arm.
At deployment, we introduce a \textbf{Latent-Aware Controller (LAC)} that interprets the learned coordination representations to modulate synchronization strength, execution asymmetry, motion smoothness, and safety constraints in real time. LAC operates at the joint-command level and remains compatible with standard robot control pipelines, without requiring force or impedance control.
Experimental results across simulation and real-world benchmarks demonstrate that Co-VLA significantly outperforms monolithic VLA baselines, achieving a 27\% success rate gain in tight-coordination tasks and more than doubling performance in out-of-distribution (OOD) real-world scenarios (from 13\% to 27\%), and reducing task completion time by up to 25\%.
\end{abstract}

\section{INTRODUCTION}

Vision–language–action (VLA) models enable robots to map visual observations and language instructions directly to continuous actions, achieving strong results in single-arm manipulation and encouraging progress in bimanual settings~\cite{rt2, openvla, pi0}. Prior works show that coordinated bimanual behaviors can emerge from end-to-end learning without explicit collaboration modeling~\cite{aloha, mobile_aloha, rdt}, suggesting that large-capacity architectures can implicitly resolve inter-arm coordination in many cases.

However, as bimanual tasks become more tightly coupled, implicit collaboration alone may be insufficient. Bimanual manipulation often requires precise temporal synchronization, asymmetric role assignment, and strict execution-level constraints such as collision avoidance and motion smoothness, spanning multiple coordination regimes with distinct structural patterns in action space. In existing VLA systems, actions for both arms are generated via direct regression from a shared representation without separating task-level coordination intent from residual execution details. This places the full burden of resolving collaboration trade-offs on the learned model, leaving synchronization strength, execution asymmetry, and safety behavior implicit and difficult to interpret or adjust at deployment time.

Our work is motivated by a key observation: \textit{collaboration in bimanual manipulation is not an action itself, but a structure over actions.} Both arms often share a common task-level intent, such as jointly transporting an object or synchronizing a handover, while requiring distinct execution strategies at the joint level. Representing bimanual actions as a monolithic vector conflates these fundamentally different sources of variation, limiting generalization and making it difficult to diagnose collaboration failures.



Our VLA system introduces explicit structure into both action generation and execution for bimanual manipulation. Our approach builds upon existing VLA frameworks and augments them with three key components. 
First, we design a Structured Action Expert (SAE) that decomposes action prediction into a shared latent and arm-specific residual latents. The shared latent represents task-level synchronization intent, while the residual latents encode asymmetric execution details for each manipulator. This structured decomposition provides a strong inductive bias for learning coordinated behaviors and enables more interpretable latent representations. In addition, we introduce task-adaptive coordination losses that enforce semantically meaningful separation between shared and residual components.
Second, we introduce a Latent-Aware Controller (LAC) that operates at deployment time to interpret the learned coordination representations. LAC modulates synchronization strength, execution asymmetry, motion smoothness, and safety constraints in real time, refining predicted joint commands before execution. This design separates learning from execution shaping and safety enforcement, allowing reliable deployment without requiring specialized hardware or low-level impedance control.
Third, we additionally explore a Co-Motion demonstration paradigm for data collection, in which both robot arms act concurrently rather than sequentially. By decomposing tasks into staged parallel motions with shared reference frames, look-ahead planning, and safe intermediate targets, Co-Motion increases the density of concurrent coordination samples in the training set, complementing SAE's structural decomposition with richer demonstration coverage.

In summary, this work makes the following contributions:

(1) We introduce \textbf{Co-VLA}, a structured action-head extension  compatible with pretrained VLA backbones. Its core, the \textbf{Structured Action Expert (SAE)}, decomposes bimanual actions into a shared coordination latent and arm-specific residual latents, paired with task-adaptive coordination losses that enforce semantically meaningful separation according to task-specific coordination patterns.

(2) We design a \textbf{Latent-Aware Controller (LAC)} that interprets the learned shared--residual representations at deployment to refine synchronization, asymmetry, and smoothness, while remaining fully compatible with standard joint-level control pipelines.

(3) We explore \textbf{Co-Motion}, a collaborative demonstration paradigm that collects concurrent bimanual trajectories through staged parallel scheduling and shared reference frames, revealing an efficiency-versus-learnability trade-off that highlights open challenges in learning from high-density coordination data.

\section{RELATED WORK}

\paragraph{VLA for Robotic Manipulation}
VLA models map multimodal inputs directly to robot actions. RT-2~\cite{rt2} first co-fine-tuned a large VLM on robotic data for emergent generalization; OpenVLA~\cite{openvla} achieved comparable results with 7B open-source parameters; Octo~\cite{octo} introduced a diffusion-based generalist policy pretrained on 800k trajectories. $\pi_0$~\cite{pi0} proposed a flow-matching VLA on a PaliGemma backbone for dexterous multi-embodiment control, extended by $\pi_{0.5}$~\cite{pi05} for open-world generalization. Diffusion Policy~\cite{diffusion_policy} and RDT-1B~\cite{rdt} further advanced continuous action modeling for bimanual tasks. While these systems show that coordinated dual-arm behavior can emerge implicitly~\cite{aloha, mobile_aloha, aloha_unleashed}, their action heads project to monolithic vectors without separating coordination intent from arm-specific execution. We retain the $\pi_0$ backbone but introduce structured decomposition via SAE at the action head.

\paragraph{Dual-Arm Coordination}
Classical bimanual control builds on operational space formulations~\cite{khatib1987}, the virtual linkage model for multi-grasp internal forces~\cite{williams1993}, and hybrid position/force coordination~\cite{uchiyama1988}. Surveys~\cite{smith2012, caccavale2016} cover impedance-based and kinematic consistency methods. Dynamical-systems approaches~\cite{mirrazavi2018} generate synchronized multi-arm motions with coordination constraints. These methods impose coordination analytically at the control level. In contrast, we introduce representation-level structure compatible with standard joint-level pipelines, bridging learning-based flexibility with structured coordination bias.

\paragraph{Structured Action Representation}
Shared--private decomposition has been used in multi-task learning~\cite{ruder2017} and multi-agent RL (e.g., QMIX~\cite{qmix}, MAPPO~\cite{mappo}) to separate common objectives from individual behaviors. More recently, Hao et al. \cite{smp2025} explored abstracting manipulation skills through mixture-of-experts in diffusion-based policies. For bimanual manipulation, IACE~\cite{iace2025} proposed inter-arm coordinated encoders for synchronization. However, these ideas have not been integrated into VLA action heads. 

\paragraph{Demonstration Paradigms for Bimanual Learning}
ALOHA~\cite{aloha} and Mobile ALOHA~\cite{mobile_aloha} collect bimanual demonstrations via puppeteering teleoperation; ALOHA Unleashed~\cite{aloha_unleashed} scaled this for dexterous tasks. RoboTwin~\cite{robotwin} and RoboTwin~2.0~\cite{robotwin2} generate scripted demonstrations via LLM-based code synthesis. In both paradigms, arm motions are often sequential or loosely coupled, underrepresenting temporal coordination. 

\paragraph{Execution Refinement}
Residual policy learning~\cite{silver2018, johannink2019} and control barrier functions~\cite{ames2019} refine learned policies for stability and safety. For bimanual systems, impedance controllers~\cite{caccavale2008, wimboeck2007} enforce coordination at the hardware level but require force sensing. Our LAC operates at the joint-command level, interpreting learned shared and residual action components to refine synchronization and smoothness without specialized hardware or backbone modifications.


\section{METHOD}
Our approach extends exisiting VLA framework by introducing structured action modeling and execution refinement for coordinated bimanual manipulation. As illustrated in Fig.~\ref{fig:system_overview}, The system comprises three components: 
\textbf{Structured Action Expert (SAE)} 
a shared--residual decomposition that explicitly separates task-level coordination intent from arm-specific execution (Sec.~\ref{sec:covla}); \textbf{Latent-Aware Controller (LAC)} interprets the learned shared and residual representations at deployment to modulate synchronization strength, motion smoothness, and safety in real time (Sec.~\ref{sec:controller}); additionally \textbf{Collaborative Motion (Co-Motion)}, an exploratory demonstration paradigm that collects concurrent bimanual trajectories in simulation, enriching the training distribution with high-density coordination samples (Sec.~\ref{sec:comotion}).

\subsection{Structured Action Expert (SAE)}\label{sec:covla}

\begin{figure}[t]
    \centering
    \includegraphics[width=0.99\linewidth]{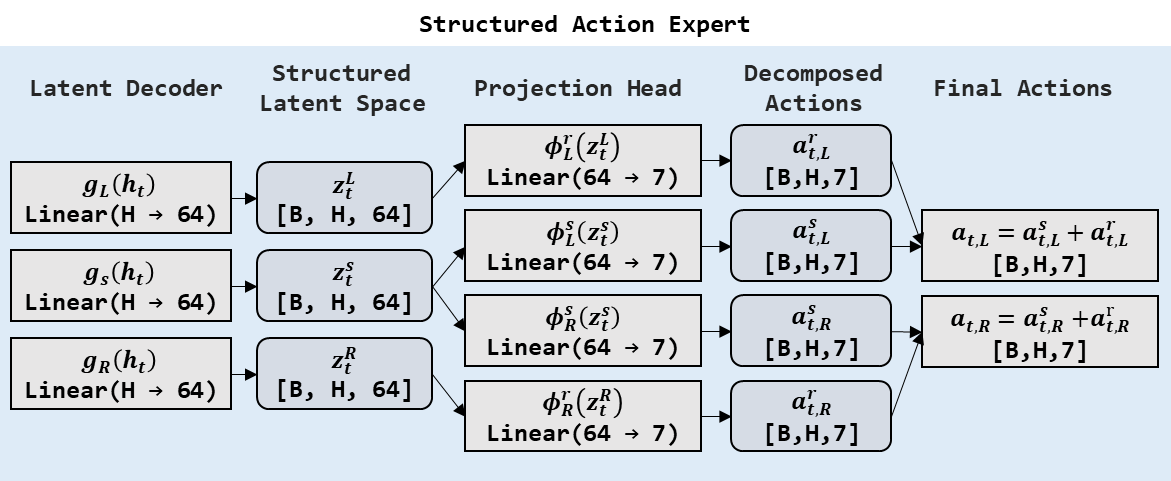}
    \caption{Structured Action Expert (SAE) architecture.}
    \label{fig:co-vla-SAE}
\end{figure}

\paragraph{Unstructured Action Generation in Standard VLAs}
In state-of-the-are VLA models such as $\pi_0$~\cite{pi0}, the action head maps transformer hidden states directly to joint-level commands. 
Given the hidden representation at time step $t$, denoted as $h_t \in \mathbb{R}^H$, the bimanual action is obtained via a single projection:
$a_t = f_{\pi_0}(h_t)$,
where $a_t = [a_{t,L}, a_{t,R}] \in \mathbb{R}^{14}$ denotes the concatenated joint commands of the left and right arms. 
This formulation treats the dual-arm action as a monolithic vector, leaving inter-arm coordination to be learned implicitly.

\paragraph{Shared--Residual Action Decomposition}

As illustrated in Fig.~\ref{fig:co-vla-SAE}, SAE replaces the unstructured projector with parallel latent decoders 
that decompose action generation into shared and residual components. 
Given the same hidden representation $h_t$, we compute three latent vectors:
\[
z_t^{s} = g_s(h_t), \qquad
z_t^{L} = g_L(h_t), \qquad
z_t^{R} = g_R(h_t),
\]
where share latent $z_t^{s} \in \mathbb{R}^{L}$ encodes shared coordination intent, left latent $z_t^{L}\in \mathbb{R}^{L}$ and right latent $z_t^{R} \in \mathbb{R}^{L}$ encode execution-specific information for each arm.

The shared latent produces common action components for each arm with separate projection heads,
\[
a_{t,L}^{s} = \phi_L^{s}(z_t^{s}), \qquad
a_{t,R}^{s} = \phi_R^{s}(z_t^{s}),
\]
while the residual latents produce residual action components for each arm again with separate projection heads,
\[
a_{t,L}^{r} = \phi_L^{r}(z_t^{L}), \qquad
a_{t,R}^{r} = \phi_R^{r}(z_t^{R}).
\]

The final joint-level commands are obtained via additive composition:
\[
a_{t,L} = a_{t,L}^{s} + a_{t,L}^{r}, \qquad
a_{t,R} = a_{t,R}^{s} + a_{t,R}^{r},
\]
where $a_{t,L}, a_{t,R} \in \mathbb{R}^{7}$ denote the predicted joint velocities for the two manipulators.
This decomposition preserves the original joint-level action interface while introducing an explicit separation between coordination structure and execution-specific adjustments.

\paragraph{Task-Adaptive Coordination Losses}
While the shared--residual decomposition provides a structural inductive bias, different bimanual tasks exhibit distinct coordination regimes. Some tasks require highly symmetric motion, others involve asymmetric role assignment, and some demand tight temporal synchronization without directional symmetry. To accommodate these variations, we introduce a modular set of auxiliary losses that shape the coordination structure in complementary ways.

\textbf{Sparse Residual Regularization.}
To encourage shared-dominant behavior in symmetric tasks, we apply an $\ell_1$ regularization on residual components:
\begin{equation}
\mathcal{L}_{\text{sparse}} = \mathbb{E}_{t} \left[ \|a_{t,L}^{r}\|_1 + \|a_{t,R}^{r}\|_1 \right].
\end{equation}
This loss biases the model toward encoding common motion in the shared latent, while allowing residual components to activate only when asymmetric adjustments are necessary.

\textbf{Shared Mean Velocity Consistency.}
To explicitly align the shared component with common motion trends, we introduce a shared-mean velocity loss:
\begin{equation}
\mathcal{L}_{\text{shared}} = \mathbb{E}_{t} \left[ \|a_{t,L}^{s} - \bar{u}_t\|_2^2 + \|a_{t,R}^{s} - \bar{u}_t\|_2^2 \right],
\end{equation}
where $\bar{u}_t = \frac{1}{2}(u_{t,L} + u_{t,R})$ is the average joint velocity. This loss reinforces the semantic interpretation of the shared latent as encoding task-level coordination intent.

\textbf{Temporal Synchronization Loss.}
Let $a_{t,L}, a_{t,R} \in \mathbb{R}^7$ denote the predicted joint velocities for the left and right arms.
We define temporal differences
$\Delta a_{t,L}=a_{t,L}-a_{t-1,L}$ and
$\Delta a_{t,R}=a_{t,R}-a_{t-1,R}$,
and their magnitudes
$m_{t,L}=\|\Delta a_{t,L}\|_2$ and
$m_{t,R}=\|\Delta a_{t,R}\|_2$.
After standardizing each magnitude sequence over time, we compute the temporal coupling as
$\mathrm{corr}_{\text{pred}} = \mathbb{E}_t \left[ \tilde m_{t,L}\tilde m_{t,R}  \right]$.
The synchronization loss is then defined as
\begin{equation}
\mathcal{L}_{\text{sync}} = 1 - \mathrm{corr}_{\text{pred}},
\end{equation}
which encourages the two arms to accelerate and decelerate synchronously regardless of motion direction.

In general, the training objective takes the form
\begin{equation}
\mathcal{L} = \mathcal{L}_{\text{action}} + \lambda \, \mathcal{L}_{\text{aux}},
\end{equation}
where $\mathcal{L}_{\text{aux}} \in 
\{\mathcal{L}_{\text{sparse}},\, \mathcal{L}_{\text{shared}},\, 
\mathcal{L}_{\text{sync}}\}$ is selected based on the dominant 
coordination pattern of the target task: 
$\mathcal{L}_{\text{sparse}}$ for near-symmetric execution, 
$\mathcal{L}_{\text{shared}}$ for asymmetric role assignment, and 
$\mathcal{L}_{\text{sync}}$ for temporally coupled motion. 
We set $\lambda = 0.001$ across all experiments. 
This task-level selection is guided by prior knowledge of each task's coordination structure; empirical validation of this design choice is provided in Experiment Section.

In practice, we observe that different tasks benefit from different auxiliary losses, reflecting variations in coordination structure. 
We interpret this as evidence that bimanual manipulation spans multiple coordination regimes, and that a modular structural bias enables targeted adaptation. 

\subsection{Latent-Aware Controller (LAC)}\label{sec:controller}

\begin{algorithm}[t]
\caption{Latent-Aware Controller (LAC)}\label{alg:controller}
\begin{algorithmic}[1]
\Require Raw action chunk $\{q_t\}_{t=1}^{H}$ where $q_t = [a_{t,L};\, a_{t,R}] \in \mathbb{R}^{14}$;
    latent components $\{a_{t,L}^s, a_{t,R}^s, a_{t,L}^r, a_{t,R}^r\}_{t=1}^{H}$
\Ensure Refined action sequence $\{\tilde{q}_t\}_{t=1}^{H}$
\State Initialize $\tilde{q}_0 \gets q_1$, \; $\alpha_0 \gets \alpha_{\mathrm{base}}$
\For{$t = 1, \dots, H$}
    \State \textcolor{gray}{\textit{\% Step 1: Energy analysis}}
    \State $E_t^{s} \gets \frac{1}{2}(\|a_{t,L}^{s}\|_2 + \|a_{t,R}^{s}\|_2)$ ,
    \State $E_t^{r} \gets \frac{1}{2}(\|a_{t,L}^{r}\|_2 + \|a_{t,R}^{r}\|_2)$
    \State $\rho_t \gets E_t^{r} \,/\, (E_t^{s} + \varepsilon)$
    \State \textcolor{gray}{\textit{\% Step 2: Opposition analysis}}
    \State $\omega_t \gets - (\langle a_{t,L}^{r},\, a_{t,R}^{r} \rangle )\,/\,(\|a_{t,L}^{r}\|_2 \, \|a_{t,R}^{r}\|_2 + \varepsilon)$
    \State \textcolor{gray}{\textit{\% Step 3: Adaptive stiffness}}
    \If{$\rho_t < \tau_{\rho}$}
    \State $\hat{\alpha}_t \gets \alpha_{\mathrm{base}} + \Delta_{\mathrm{macro}}$ \Comment{Macro-dominant}
\ElsIf{$\rho_t \geq \tau_{\rho} \;\wedge\; \omega_t > \tau_{\omega}$}
    \State $\hat{\alpha}_t \gets \alpha_{\mathrm{base}} + \Delta_{\mathrm{prec}}$ \Comment{Precision-critical}
\Else
    \State $\hat{\alpha}_t \gets \alpha_{\mathrm{base}} - \Delta_{\mathrm{noise}}$ \Comment{Noise suppression}
\EndIf
    \State $\hat{\alpha}_t \gets \mathrm{clip}(\hat{\alpha}_t,\; \alpha_{\min},\; \alpha_{\max})$
    \State $\alpha_t \gets (1 - \beta)\,\alpha_{t-1} + \beta\,\hat{\alpha}_t$
    \State \textcolor{gray}{\textit{\% Step 4: Low-pass refinement}}
    \State $\tilde{q}_t \gets (1 - \alpha_t)\,\tilde{q}_{t-1} + \alpha_t\,q_t$
\EndFor
\end{algorithmic}
\end{algorithm}

SAE produces joint-level action predictions together with the shared and residual latent components. While these predictions already encode meaningful bimanual structure, direct execution can still suffer from high-frequency jitter, overreaction to small residual signals, or insufficient protection of precision-critical motions. The Latent-Aware Controller (LAC) addresses these issues by interpreting the shared--residual decomposition at deployment time to adaptively refine joint commands before they are sent to the robot.

Recall that the predicted actions are decomposed as $a_{t,L} = a_{t,L}^{s} + a_{t,L}^{r}, 
a_{t,R} = a_{t,R}^{s} + a_{t,R}^{r},$
where the shared components $a^{s}$ represent task-level collaboration (macro motion), and the residual components $a^{r}$ represent residual execution details (micro adjustment).

Empirically, we observe that in most time steps the shared components dominate the action magnitude, while the residual components are relatively small but critical for precise collaboration. This motivates a macro--micro control strategy in which shared actions are transmitted with high fidelity, while residual actions are selectively protected or suppressed depending on their semantic relevance.

\paragraph{Energy-Based Micro--Macro Analysis}
At each time step $t$, we compute the average magnitudes of the shared and residual components:
\[
E_t^{s} = \frac{1}{2}\left(\|a_{t,L}^{s}\|_2 + \|a_{t,R}^{s}\|_2\right), \
E_t^{r} = \frac{1}{2}\left(\|a_{t,L}^{r}\|_2 + \|a_{t,R}^{r}\|_2\right).
\]
We define the micro-motion ratio as
\[
\rho_t = \frac{E_t^{r}}{E_t^{s} + \varepsilon},
\]
where $\varepsilon$ is a small constant for numerical stability. This ratio characterizes the relative importance of residual adjustments at the current time step.

\paragraph{Residual Opposition and Collaboration Signal}
To distinguish meaningful asymmetric collaboration from noise, we further analyze the directional relationship between the residual components. We compute the cosine similarity
\[
\mathrm{cos}_t =
\frac{\langle a_{t,L}^{r}, a_{t,R}^{r} \rangle}
{\|a_{t,L}^{r}\|_2 \, \|a_{t,R}^{r}\|_2 + \varepsilon},
\]
and define an opposition score $\omega_t = - \mathrm{cos}_t$.
A high opposition score indicates that the two arms apply residual adjustments in opposing directions, which commonly arises in coordinated behaviors such as stabilizing, holding, or fine alignment. In contrast, low or inconsistent opposition suggests residual noise or jitter.

\paragraph{Adaptive Stiffness Modulation}
Based on the micro-motion ratio $\rho_t$ and opposition score $\omega_t$, the controller computes an adaptive execution stiffness $\hat{\alpha}_t$ around a base stiffness $\alpha_{\mathrm{base}}$. The controller follows three regimes:
\begin{itemize}
    \item \textbf{Macro-dominant motion:} If $\rho_t$ is below a small threshold $\tau_{\rho}$, the action is dominated by shared motion. The controller increases stiffness by a margin $\Delta_{\mathrm{macro}}$ ($\hat{\alpha}_t = \alpha_{\mathrm{base}} + \Delta_{\mathrm{macro}}$) to allow fast, direct execution.
    \item \textbf{Precision-critical micro adjustment:} If $\rho_t \geq \tau_{\rho}$ and $\omega_t$ exceeds a collaboration threshold $\tau_{\omega}$, residual actions are interpreted as meaningful micro adjustments. The stiffness is explicitly increased ($\hat{\alpha}_t = \alpha_{\mathrm{base}} + \Delta_{\mathrm{prec}}$) to preserve these delicate opposite-direction alignments.
    \item \textbf{Noise suppression:} Otherwise, the residual energy is considered uncoordinated high-frequency jitter and stiffness is reduced ($\hat{\alpha}_t = \alpha_{\mathrm{base}} - \Delta_{\mathrm{noise}}$) to suppress it.
\end{itemize}

To avoid abrupt changes, the target stiffness $\hat{\alpha}_t$ is first bounded within $[\alpha_{\min}, \alpha_{\max}]$, and then temporally smoothed:
\[
\alpha_t = (1-\beta)\,\alpha_{t-1} + \beta\,\hat{\alpha}_t,
\]
where $\beta \in (0,1)$ controls the smoothing rate.

\paragraph{Joint-Level Action Refinement}
Given the raw joint command $q_t = [a_{t,L};\, a_{t,R}]$ predicted by the VLA model, the controller applies a first-order low-pass refinement:
\[
\tilde{q}_t = (1-\alpha_t)\,\tilde{q}_{t-1} + \alpha_t\,q_t,
\]
where $\tilde{q}_t$ denotes the refined joint command sent to the robot. This formulation ensures smooth execution while adaptively preserving collaboration-critical motion components.

LAC does not introduce additional learning modules and does not modify the trained policy. Instead, it interprets the semantic structure already present in SAE's action decomposition to shape execution behavior. By explicitly distinguishing macro collaboration from micro adjustment and protecting meaningful residual signals, LAC improves motion smoothness, stability, and safety without requiring force sensing or impedance control.

\begin{figure}[t]
    \centering
    \includegraphics[width=\linewidth]{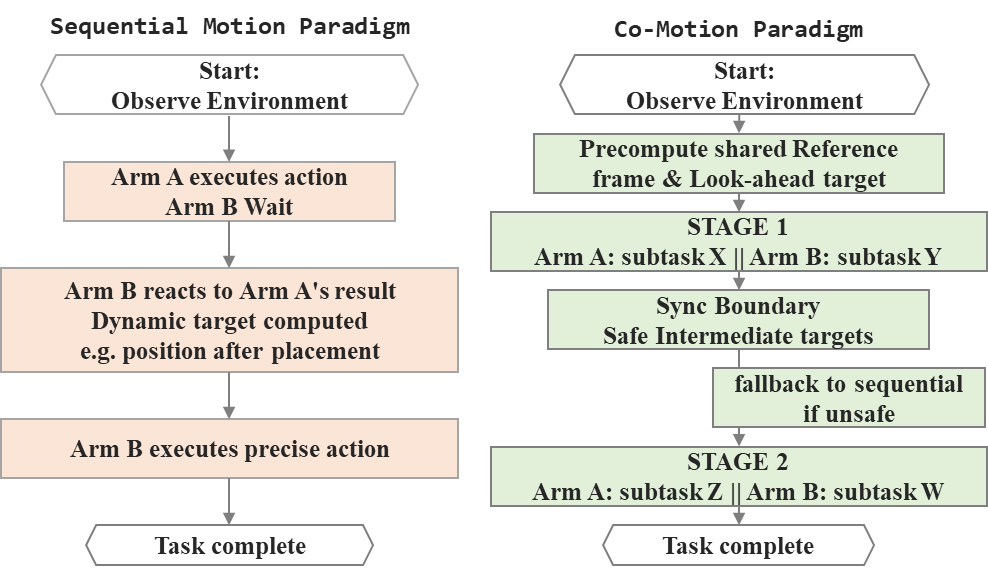}
    \caption{Visualization of sequential motion paradigm (left) and our designed collaborative motion paradigm (right). }
    \label{fig:Co-Motion}
\end{figure}

\subsection{Co-Motion: Collaborative Motion Paradigm}\label{sec:comotion}

Standard bimanual demonstration pipelines typically execute arm motions sequentially or with loose synchronization, underrepresenting the temporal coordination and role coupling that characterize real collaborative manipulation. Co-Motion is an exploratory demonstration paradigm that increases the density of concurrent bimanual coordination in the training set. It is implemented by restructuring the task-level scheduling logic within the RoboTwin~2.0 code-generation pipeline; the underlying motion planner (cuRobo) and primitive interfaces remain unchanged.

As illustrated in Fig.~\ref{fig:Co-Motion}, Co-Motion decomposes each task into a sequence of stages with explicit synchronization boundaries. Within each stage, non-conflicting subtasks are dispatched to both arms in parallel, reducing idle time and capturing natural temporal overlap. To make this parallelization practical, Co-Motion adopts three supporting mechanisms: (i)~\emph{shared reference frames}, namely common spatial anchors (e.g., a handover midpoint or goal pose) that both arms reference, decoupling their planning and stabilizing demonstrations across trials; (ii)~\emph{look-ahead precomputation} of near-future interaction targets, which eliminates stop-and-plan interruptions and yields smoother trajectories; and (iii)~\emph{safe intermediate targets} with explicit clearance margins that act as buffers between stages, with graceful fallback to sequential execution when parallel dispatch is infeasible.

Co-Motion is a complementary data strategy rather than a prerequisite: SAE and LAC already deliver consistent improvements on standard sequential demonstrations (cf.\ Table~\ref{tab:robotwin-sr}).  Co-Motion provides an additional axis for enriching the training distribution, 
revealing an efficiency-versus-learnability trade-off that warrants further investigation.

\section{Experiment}

\begin{figure}[t]
    \centering
    \small{\textbf{RoboTwin tasks}}\vspace{0.1cm}\\
    \includegraphics[width=\linewidth]{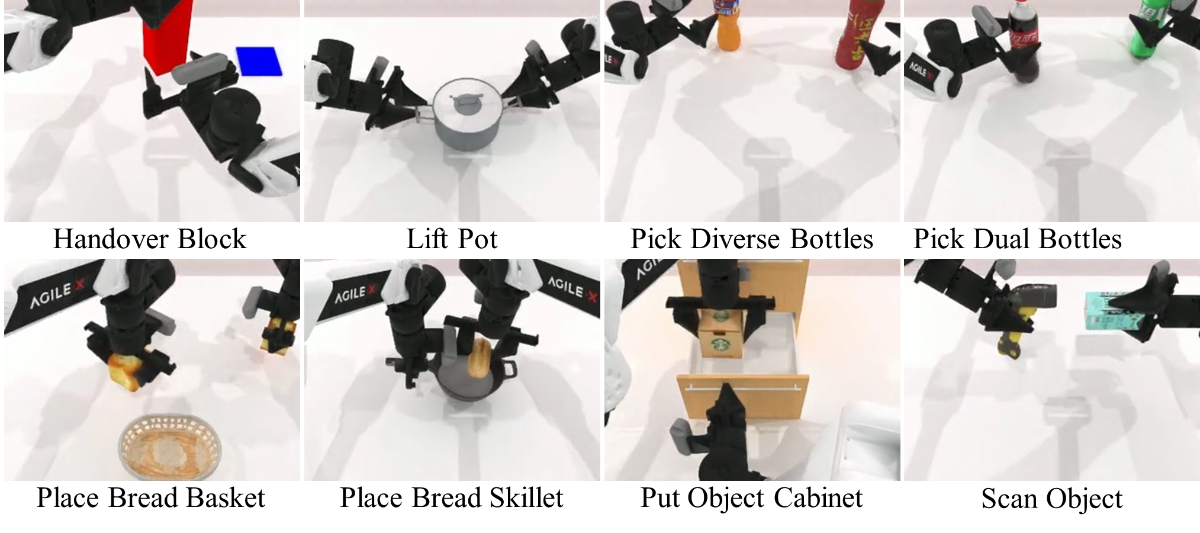}
    \small{\textbf{Real-World tasks}}\vspace{0.1cm} \\ 
    \includegraphics[width=0.99\linewidth]{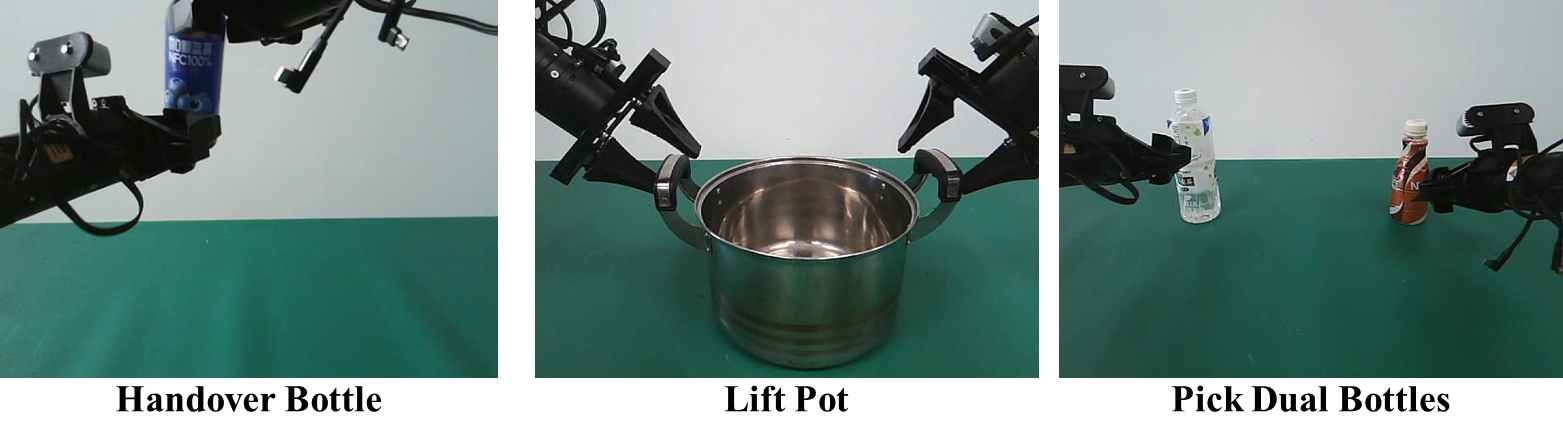}
    \caption{Evaluation tasks. \textit{Top:} Bimanual tasks in RoboTwin~2.0 simulation (Aloha-AgileX robot). \textit{Bottom:} Real-world tasks (AgileX Cobot Magic robot).}
    \label{fig:all_tasks}
\end{figure}

\subsection{Training Pipeline}
At training stage, we adopt a two-phase strategy. In Phase~1 (warm-up), we freeze the pretrained backbone and train only the newly introduced SAE layers, including the shared and residual action projections as well as the share-to-arm routing modules, for 1{,}000 steps with a peak learning rate of $5 \times 10^{-5}$. This allows the new parameters to reach a reasonable initialization without disrupting the pretrained representations. In Phase~2 (full fine-tuning), we resume from the Phase~1 checkpoint, unfreeze all parameters, and continue training for 30{,}000 steps with a reduced peak learning rate of $2.5 \times 10^{-5}$ that decays to $2.5 \times 10^{-6}$. Both phases use a batch size of 32 distributed across 4 GPUs with FSDP.

\subsection{Simulation Evaluation on RoboTwin 2.0}
In simulation, we conduct experiments on the RoboTwin~2.0 benchmark, which offers over 50 dual-arm manipulation tasks. From this pool, we deliberately select a subset of tasks that explicitly require concurrent bimanual motion, as illustrated in Fig.~\ref{fig:all_tasks}. The selected tasks still span diverse coordination patterns, including symmetric lifting, asymmetric role assignment, and temporally coupled handovers. For each task, we collect 1{,}000 successful demonstrations under clean (non-randomized) scene configurations, fine-tune a separate model per task, and evaluate over 100 rollouts under both Easy and Hard settings with increasing levels of scene randomization.

\subsubsection{Co-VLA on RoboTwin Tasks}
We first compare Co-VLA with state-of-the-arts methods on bimanual manipulation tasks in RoboTwin~2.0. Both models are trained on the same demonstration data and evaluated under identical conditions. Performance is measured by the average task success rate over 100 rollouts.

\begin{table}[t]
    \centering
    \caption{Success rate (\%) on RoboTwin~2.0 tasks over 100 rollouts. Higher is better.}\label{tab:robotwin-sr}
    \resizebox{\linewidth}{!}{%
    \begin{tabular}{l*{6}{c}}
    \toprule
          & \multicolumn{2}{c}{$\pi_{0.5}$} & \multicolumn{2}{c}{$\pi_0$} & \multicolumn{2}{c}{Co-VLA (Ours)} \\ 
          \cmidrule(lr){2-3} \cmidrule(lr){4-5} \cmidrule(lr){6-7}
          & Easy & Hard & Easy & Hard & Easy & Hard \\ 
    \midrule
        Handover Block & 44\% & 10\% & 64\% & 7\% & \textbf{91}\% & \textbf{12}\% \\
        Lift Pot & 100\% & 60\% & 100\% & 63\% & 100\% & \textbf{65}\% \\
        Pick Diverse Bottles & 95\% & 18\% & 91\% & 13\% & \textbf{95}\% & 16\% \\
        Pick Dual Bottles & 100\% & \textbf{27}\% & 100\% & 16\% & 100\% & 18\% \\
        Place Bread Basket & \textbf{92}\% & 28\% & 70\% & 46\% & 69\% & \textbf{48}\% \\
        Place Bread Skillet & 38\% & 7\% & 63\% & 8\% & \textbf{70}\% & \textbf{8}\% \\
        Put Object Cabinet & 70\% & \textbf{19}\% & 79\% & 8\% & \textbf{81}\% & 5\% \\
        Scan Object & 45\% & \textbf{6}\% & 41\% & 3\% & \textbf{50}\% & 2\% \\
     \midrule
     Average & 73\% & 21.9\% & 76\% & 21\% & \textbf{82}\% & \textbf{22}\% \\
    \bottomrule
    \end{tabular}}
\end{table}
As shown in Table~\ref{tab:robotwin-sr}, Co-VLA consistently matches or improves upon both $\pi_0$ and $\pi_{0.5}$ across tasks, raising the average Easy-setting success rate from 76\% ($\pi_0$) and 73\% ($\pi_{0.5}$) to 82\%. The largest improvement appears in Handover Block (64\%$\to$91\% over $\pi_0$; 44\%$\to$91\% over $\pi_{0.5}$), a task requiring tight inter-arm coordination, suggesting that explicit shared--residual decomposition particularly benefits role-coupled manipulation. Notably, $\pi_{0.5}$ does not consistently outperform $\pi_0$ on these bimanual-specific tasks, indicating that backbone capacity alone is insufficient to resolve inter-arm coordination without structural inductive bias.

\subsubsection{Effect of Co-Motion Demonstrations}\label{sec:exp-comotion}

We analyze Co-Motion along two axes \emph{execution efficiency} and \emph{learnability} to characterize the trade-off introduced by concurrent demonstration collection.

Not all bimanual tasks  admit a Co-Motion variant: concurrent arm dispatch requires that parallel subtasks are spatially non-conflicting and that the underlying motion planner (cuRobo) maintains a sufficiently high success rate under the tighter timing constraints. Of the selected tasks, four satisfy both criteria: Handover Block, Scan Object, Place Bread Skillet, and Put Object Cabinet. 

\paragraph{Efficiency}
We first measure the average time to generate 1{,}000 successful demonstrations per task. As shown in Table~\ref{tab:demo-time}, Co-Motion reduces demonstration time by 10--25\% across all four tasks, confirming that parallel scheduling and shared anchors improve data collection throughput without sacrificing success rate.

\begin{table}[t]
\centering
\caption{Average time (seconds) to generate 1{,}000 successful demonstrations. Lower is better.}\label{tab:demo-time}
\begin{tabular}{l*{4}{c}}
\toprule
 & \multicolumn{4}{c}{Task} \\
\cmidrule(lr){2-5}
Method & \makecell{Handover\\Block} & \makecell{Scan\\Object} & \makecell{Place Bread\\Skillet} & \makecell{Put Object\\Cabinet} \\
\midrule
Sequential & 9.47 & 5.65 & 5.51 & 8.95 \\
Co-Motion & \textbf{8.49} & \textbf{4.52} & \textbf{4.59} & \textbf{6.45} \\
\bottomrule
\end{tabular}
\end{table}

This efficiency advantage persists at inference time. Table~\ref{tab:inference-time} reports the average completion time of successful rollouts during model execution. Models trained on Co-Motion data consistently produce faster task completions, indicating that the concurrent motion patterns transfer to more temporally compact execution behavior.

\begin{table}[t]
    \centering
    \caption{Average completion time (seconds) for 50 successful rollouts during evaluation. Lower is better.}\label{tab:inference-time}
    \setlength{\tabcolsep}{4pt}
    \begin{tabular}{l*{4}{c}}
    \toprule
           & \multicolumn{2}{c}{RoboTwin} & \multicolumn{2}{c}{Co-Motion} \\ 
          \cmidrule(lr){2-3} \cmidrule(lr){4-5}
           & $\pi_0$ & Co-VLA & $\pi_0$ & Co-VLA \\ 
    \midrule
        Handover Block & 28.39 & 28.56 & 32.87 & \textbf{25.38} \\
        Place Bread Skillet & 15.02 & 15.28 & 12.79 & \textbf{12.35} \\
        Put Object Cabinet & 26.96 & 28.69 & \textbf{20.87} & 21.08 \\ 
        Scan Object & 15.94 & 19.00 & \textbf{12.57} & 13.92 \\
    \midrule
        Average & 21.58	& 22.88	& 19.78	& \textbf{18.18} \\
    \bottomrule
    \end{tabular}
\end{table}

\paragraph{Learnability}
Despite the efficiency gains, training on Co-Motion data proves substantially more challenging than training on standard RoboTwin demonstrations. As shown in Table~\ref{tab:comotion-sr}, 
the average Easy-setting success rate drops from 76\% to 50\% for $\pi_0$, and from 82\% to 56\% for Co-VLA.
This drop is consistent across architectures, indicating that the difficulty lies in the data distribution itself rather than a specific model limitation. Concurrent trajectories introduce tighter temporal coupling and more complex inter-arm dependencies, raising the learning difficulty relative to sequential demonstrations. Notably, Co-VLA still outperforms $\pi_0$ under Co-Motion training (56\% vs.\ 50\%), suggesting that SAE's shared--residual decomposition helps the model better absorb high-density coordination signals even when overall learning is harder.
These results reveal a clear \emph{efficiency-vs-learnability} trade-off: Co-Motion produces more compact and coordination-rich demonstrations, but current VLA architectures have difficulty fully exploiting them. 

\begin{table}[t]
    \centering
    \caption{Success rate (\%) when trained with Co-Motion demonstrations. Higher is better.}\label{tab:comotion-sr}
    \begin{tabular}{l*{4}{c}}
    \toprule
          & \multicolumn{2}{c}{$\pi_0$} & \multicolumn{2}{c}{Co-VLA (Ours)} \\ 
          \cmidrule(lr){2-3} \cmidrule(lr){4-5}
          & Easy & Hard & Easy & Hard \\ 
    \midrule
        Handover Block & 15\% & 8\% & \textbf{38\%} & 8\% \\
        Place Bread Skillet & 60\% & 3\% & \textbf{62\%} & \textbf{8\%} \\
        Put Object Cabinet & \textbf{83\%} & 45\% & 70\% & \textbf{47\%} \\
        Scan Object & 43\% & \textbf{8\%} & \textbf{53\%} & 3\% \\
    \midrule
     Average & 50\% & 16\% & \textbf{56\%} & \textbf{17\%} \\
    \bottomrule
    \end{tabular}
\end{table}

\subsection{Real-World Evaluation}

\begin{figure}[t]
    \centering
    \includegraphics[width=.99\linewidth]{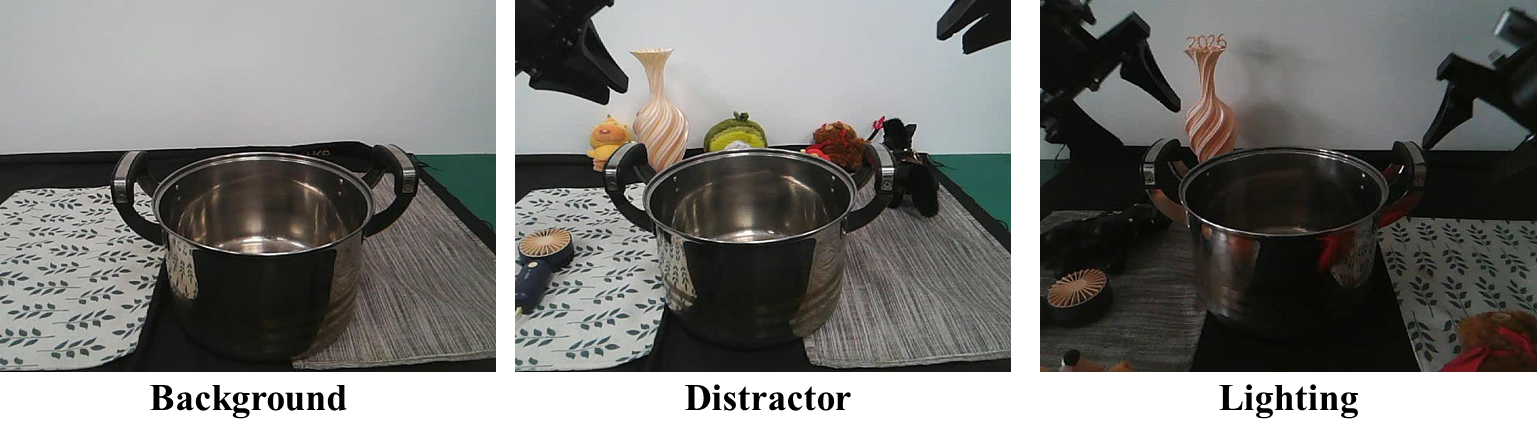}
    \vspace{-.6cm}
    \caption{Examples of out-of-distribution conditions for real-world evaluation on the Lift Pot task, including cluttered backgrounds, distractor objects, and low-lighting.}\label{fig:real_ood}    
\end{figure}

For real-world evaluation, we deploy the models on an AgileX Cobot Magic dual-arm robot equipped with wrist-mounted cameras. For each task, we collect 50 human demonstrations via teleoperation with randomized object positions and poses. In-distribution (ID) evaluation follows the same randomization protocol. 
Out-of-distribution (OOD) evaluation additionally introduces cluttered backgrounds, distractor objects, and low-lighting conditions to assess robustness beyond the training distribution, as illustrated in Fig.~\ref{fig:real_ood}.

For the Latent-Aware Controller (Algorithm~\ref{alg:controller}), we empirically configure the modulation parameters as follows: $\alpha_{\mathrm{base}}=0.4$, 
$\Delta_{\mathrm{macro}}=0.4$, 
$\Delta_{\mathrm{prec}}=0.3$, 
$\Delta_{\mathrm{noise}}=0.1$, 
$\tau_{\rho}=0.01$, 
$\tau_{\omega}=0.3$, 
$\beta=0.2$, 
and bounds $[\alpha_{\min},\alpha_{\max}]=[0.05,0.95]$. 
These parameters are kept fixed across all real-world tasks.
We mainly compare three variants: $\pi_0$, Co-VLA without LAC, and Co-VLA with LAC enabled.

\begin{table}[t]
\centering
\caption{Real-world experiment results over 30 rollouts for each evaluation under ID and OOD settings.}
\label{tab:realworld}
\begin{tabular}{lcccccc}
\toprule
 & \multicolumn{2}{c}{Handover} & \multicolumn{2}{c}{Pick Bottles} & \multicolumn{2}{c}{Lift Pot} \\
Method & ID & OOD & ID & OOD & ID & OOD \\
\midrule
$\pi_0$ & 63\% & 13\% & 57\% & 37\% & 43\% & 33\% \\
Co-VLA$_{\text{noLAC}}$ & 57\% & 17\% & 57\% & \textbf{50\%} & 63\% & 27\% \\
Co-VLA$_{\text{LAC}}$ & \textbf{73\%} & \textbf{27\%} & \textbf{67\%} & 47\% & \textbf{67\%} & \textbf{37\%} \\
\bottomrule
\end{tabular}
\end{table}
As shown in Table~\ref{tab:realworld}, the full system 
(Co-VLA$_{\text{LAC}}$) achieves the highest success rate in five out 
of six settings, with the largest gains on coordination-intensive tasks: 
Handover ID improves from 63\% ($\pi_0$) to 73\%, and Lift Pot ID from 
43\% to 67\%.

Comparing Co-VLA$_{\text{noLAC}}$ with $\pi_0$ isolates the effect of 
SAE alone. The results are task-dependent: SAE yields a notable 
improvement on Lift Pot (43\%$\to$63\% ID) but a slight decrease on 
Handover (63\%$\to$57\% ID), suggesting that structured action 
decomposition alone does not uniformly translate to higher success 
without execution-time refinement. Enabling LAC consistently recovers 
and surpasses $\pi_0$ across all ID settings, indicating that the 
combination of structured representation and latent-aware control is 
more robust than either component in isolation.

Under OOD conditions, Co-VLA$_{\text{LAC}}$ improves over $\pi_0$ on 
all three tasks (e.g., Handover 13\%$\to$27\%, Lift 33\%$\to$37\%). 
One exception is Pick Bottles OOD, where Co-VLA$_{\text{noLAC}}$ 
achieves the highest rate (50\%) while LAC slightly reduces it to 47\%, 
possibly because LAC's noise suppression filters out residual 
adjustments that happen to be beneficial under distribution shift for 
this particular task. Overall, these results highlight that SAE and LAC 
are complementary: SAE provides the representational structure, while 
LAC translates it into reliable execution.

\subsection{In-Depth Analysis}

\begin{figure}[t]
    \centering
    \includegraphics[width=0.49\linewidth]{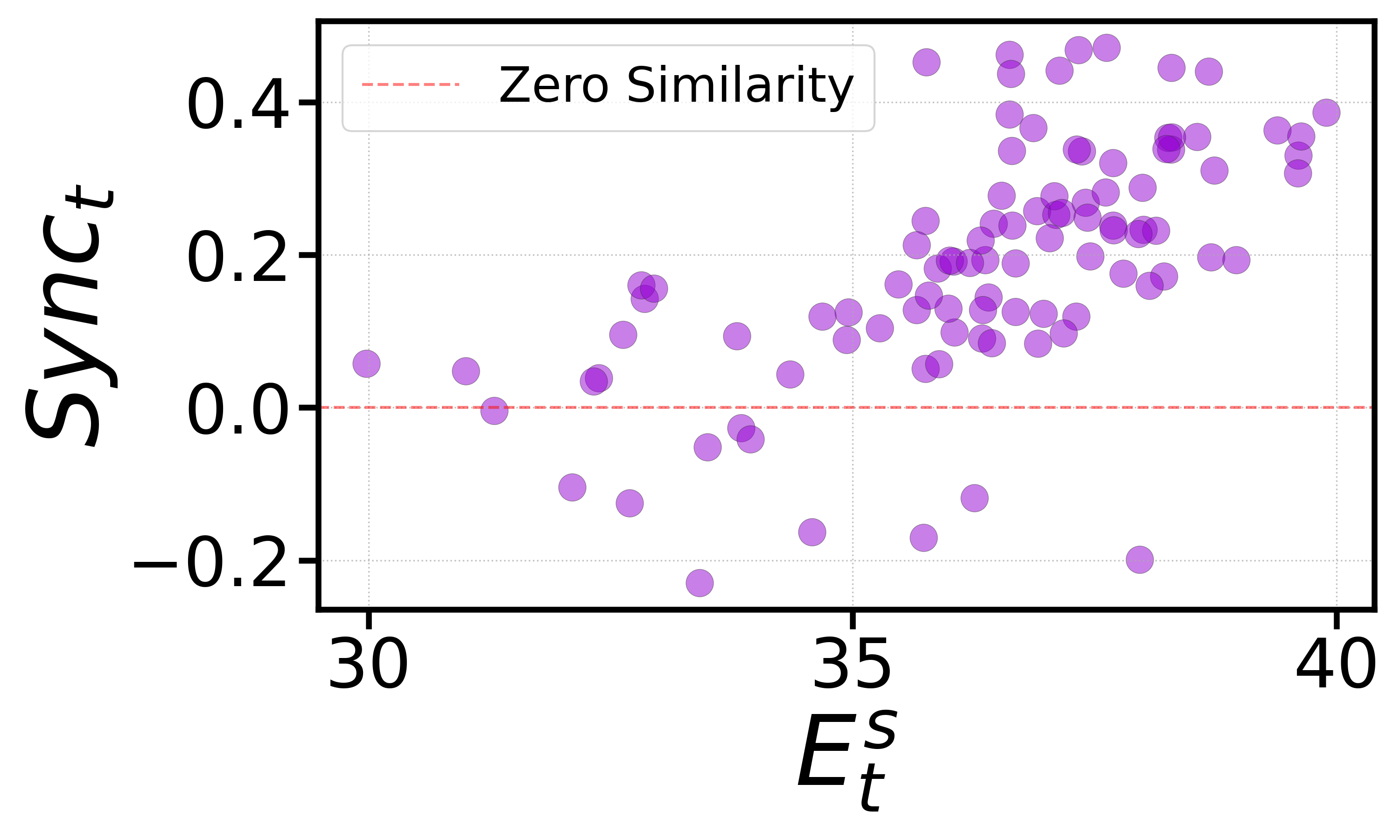}
    \includegraphics[width=0.49\linewidth]{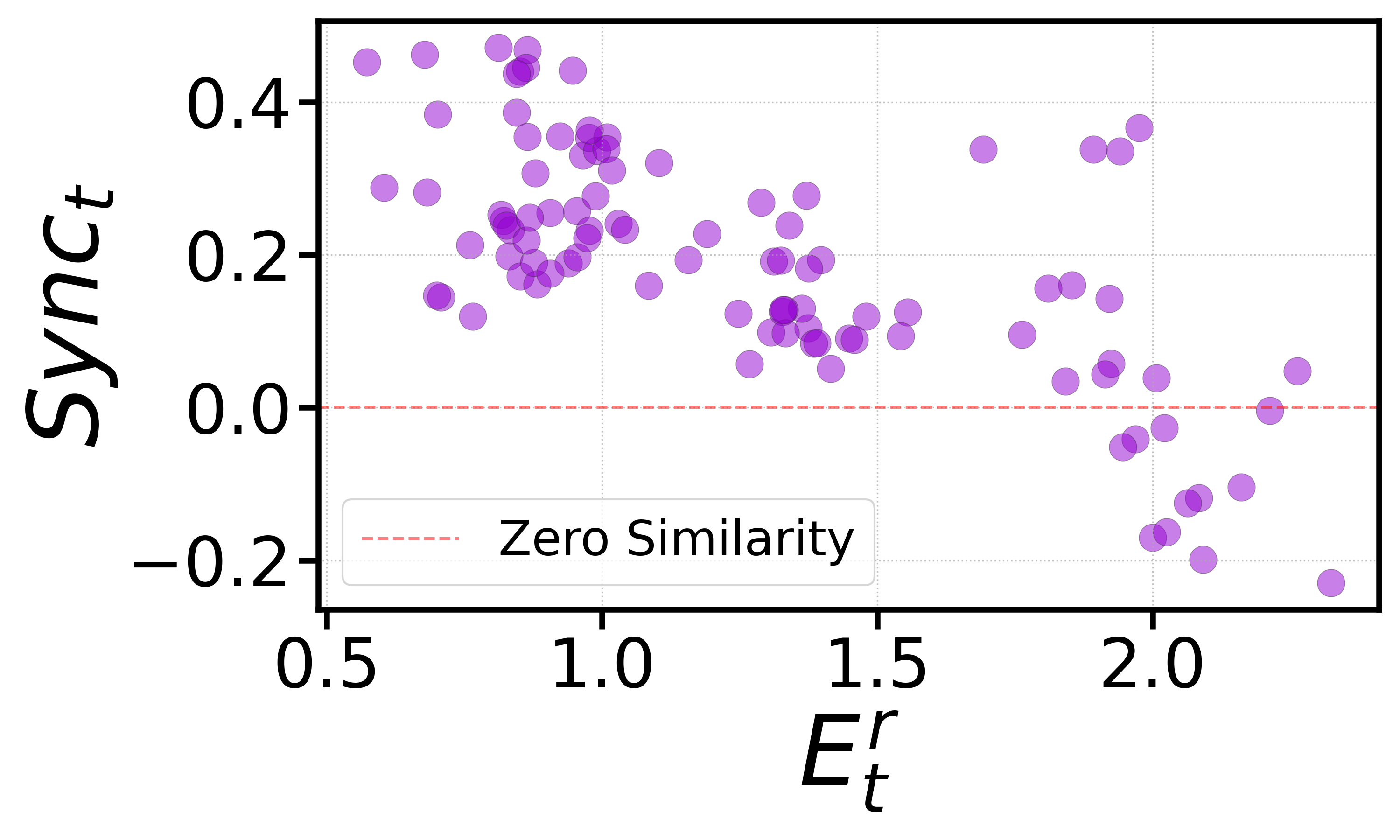}
    \vspace{-.6cm}    
    \caption{Latent--behavior alignment on the Pick Bottles task. \textit{Left:} shared energy $E_t^s$ vs.\ inter-arm synchronization $\mathrm{Sync}_t$ (positive correlation). \textit{Right:} residual energy $E_t^r$ vs.\ $\mathrm{Sync}_t$ (negative correlation).}\label{fig:latent}
\end{figure}

\paragraph{Latent–Behavior Alignment}
\textbf{Question:} \textit{Does the structured latent decomposition in Co-VLA translate into behaviorally meaningful coordination?}
We analyze how the magnitude of the shared and residual action components relates to the actual synchronization between the two arms at inference time. 
We compute $\mathrm{Sync}_t$ the cosine similarity between the final composed joint actions $a_{t,L}$ and $a_{t,R}$.
We visualize $\mathrm{Sync}_t$ against $E_t^{s}$ and $E_t^{r}$ across rollouts in Fig.~\ref{fig:latent}.

The results reveal a consistent positive correlation between 
synchronization and shared energy (Fig.~\ref{fig:latent} Left), and a 
negative correlation with residual energy (Fig.~\ref{fig:latent} Right). 
This confirms that the shared latent acts as a coordination driver 
promoting aligned motion, while the residual components encode 
arm-specific deviations---indicating that the learned decomposition 
captures interpretable coordination structure rather than merely 
partitioning internal representations.

\paragraph{Action Head Ablation and Structural Priors}
\textbf{Question:}
\textit{How do different auxiliary structural priors influence performance across tasks? }
We perform a controlled ablation study by adding each auxiliary loss individually on top of the primary objective $\mathcal{L}_{\text{action}}$. Due to computational constraints, we restrict our evaluation to single-loss variants rather than exploring all combinations.

Table~\ref{tab:loss} reports results on representative tasks from three coordination patterns. The performance gains are task-dependent. On the role-asymmetric Put Object Cabinet task, adding shared consistency produces the largest improvement (49\% $\rightarrow$ 70\%), suggesting that stabilizing common motion trends is particularly beneficial when asymmetric execution must be coordinated. On the temporally coupled Place Bread Skillet task, synchronization loss yields the highest performance (51\% $\rightarrow$ 62\%), indicating that enforcing aligned motion timing is critical in such scenarios.

\begin{table}[t]
    \centering
    \caption{Auxiliary loss ablation on representative tasks from three coordination regimes. All models use $\mathcal{L}_{\text{action}}$ as the primary objective. Each row adds one auxiliary loss. Success rate (\%) over 100 rollouts on Easy Setting. Higher is better.}
    \setlength{\tabcolsep}{4pt}  
    \begin{tabular}{lccc}
    \toprule
    Aux. Loss & Lift (Sym.) & Cabinet (Asym.) & Skillet (Temp.) \\
    \midrule
    None & 99\% & 49\% & 51\% \\
    $+\;\mathcal{L}_{\text{sparse}}$ & 100\% & 51\% & 55\% \\
    $+\;\mathcal{L}_{\text{shared}}$ & 100\% & \textbf{72\%}  & 55\% \\
    $+\;\mathcal{L}_{\text{sync}}$   & 100\% & 70\% & \textbf{62\%} \\
    \bottomrule
    \end{tabular}\label{tab:loss}
\end{table}


\paragraph{Effect of LAC}
\textbf{Question.}  
\textit{What is the role of LAC during execution, and how does it affect performance and stability?}
To isolate the effect of deployment-time execution refinement, we conduct an ablation study on the real-world Pick Bottles task, comparing Co-VLA with LAC enabled, without LAC, and against a standard Exponential Moving Average (EMA) baseline. All other components, including trained policy and raw action predictions, remain identical.

\begin{figure}[t]
    \centering
    \includegraphics[width=.9\linewidth]{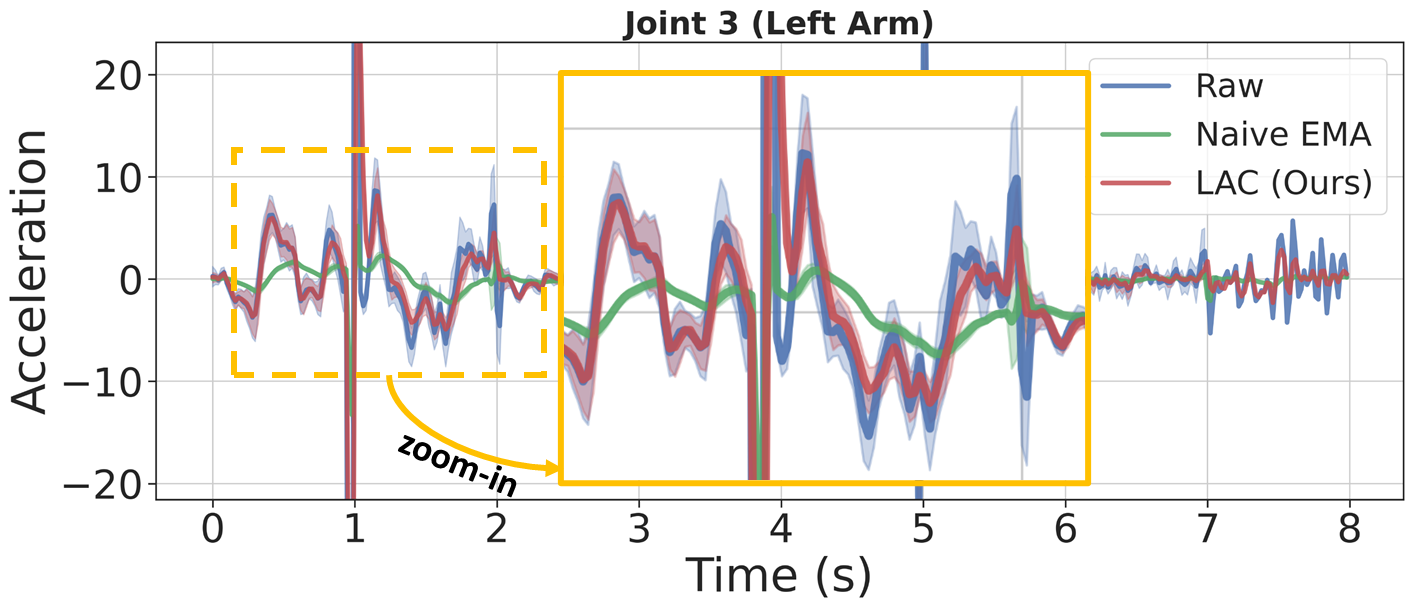}
    \caption{Effect of LAC on joint trajectories. LAC produces smoother trajectories with fewer acceleration peaks and lower cross-rollout variance while preserving task-critical motion patterns. Naive EMA filtering yields the smoothest curves but over-suppresses precision-critical residual adjustments, leading to reduced task success.}\label{fig:controller}
\end{figure}

Empirically, while the naive EMA filter produces the absolutely smoothest joint trajectories  (Fig.~\ref{fig:controller}), it suffers a notable drop in task success (Pick Bottles at 60\% compared to LAC enabled at 67\%). Qualitative inspection reveals that EMA's uniform low-pass filtering introduces unavoidable phase lag and ``over-smooths'' the trajectory. This inadvertently washes out the precision-critical micro-adjustments required for the tight physical coupling inherent in bimanual picking. 

LAC overcomes this trade-off by explicitly interpreting the semantic structure of SAE's action decomposition. Macro collaboration components are transmitted with high fidelity to preserve synchronized motion, while micro residual components are selectively protected or suppressed based on their relative magnitude and opposition structure. 
This allows LAC to filter high-frequency noise while preserving task-critical residual adjustments that EMA blindly eliminates. Since LAC operates entirely at execution time without modifying the learned policy, it highlights the complementary roles of SAE and LAC.

\section{CONCLUSION}

In this paper, we presented Co-VLA, a coordination-aware bimanual manipulation framework that introduces explicit structural priors into VLA models. Specifically, we instantiate our method on a state-of-the-art vision-language backbone by replacing its monolithic action head with a Structured Action Expert (SAE) that decomposes coordination into shared and residual latents, and integrating this with a Latent-Aware Controller (LAC) at deployment, Co-VLA achieves superior task success, execution smoothness, and latent interpretability.

Several critical insights emerged from our study. 
First, although the shared--residual decomposition is 
architecture-agnostic, validating SAE and LAC on other 
continuous-action paradigms (e.g., Diffusion Policy) remains future work.
Second, while the Co-Motion paradigm enriches the training distribution with concurrent samples, it results in a performance trade-off where success rates drop compared to sequential demonstrations. This suggests that concurrent bimanual trajectories, characterized by tighter temporal coupling, impose significantly higher demands on the spatio-temporal reasoning of VLA backbones.
Finally, due to computational constraints, our auxiliary loss ablation was limited to single-loss variants selected per task.  
Defining computable task-level coordination descriptors, such as inter-arm velocity correlation or role asymmetry indices, to enable automatic loss routing, and exploring joint activation of multiple losses with adaptive weighting, are promising directions for scaling the framework to diverse task distributions without manual loss selection.

\addtolength{\textheight}{-12cm}   



\end{document}